\documentclass[conference]{IEEEtran}
\IEEEoverridecommandlockouts
\usepackage{cite}
\usepackage{amsmath,amssymb,amsfonts}
\usepackage{algorithmic}
\usepackage{graphicx}
\usepackage{textcomp}
\usepackage{xcolor}
\usepackage{subcaption}

\usepackage{fancyhdr}

\def\BibTeX{{\rm B\kern-.05em{\sc i\kern-.025em b}\kern-.08em
    T\kern-.1667em\lower.7ex\hbox{E}\kern-.125emX}}
\begin{document}

\title{Quantitative Hardness Assessment with Vision-based Tactile Sensing for Fruit Classification and Grasping*\\

\thanks{The author extends sincere appreciation to The Hong Kong Center for Construction Robotics (InnoHK center supported by the Hong Kong ITC, InnoHK-HKCRC) for the funding support.}
}

\author{\IEEEauthorblockN{1\textsuperscript{st} Zhongyuan LIAO}
\IEEEauthorblockA{\textit{Hong Kong Center for Construction Robotics} \\
\textit{The Hong Kong University of Science and Technology}\\
Hong Kong, China \\
liaozy@ust.hk}
\and
\IEEEauthorblockN{2\textsuperscript{nd}Yipai DU}
\IEEEauthorblockA{\textit{Hong Kong Center for Construction Robotics} \\
\textit{The Hong Kong University of Science and Technology}\\
Hong Kong, China \\
yipai.du@outlook.com}
\and
\IEEEauthorblockN{3\textsuperscript{rd} Jianghua Duan }
\IEEEauthorblockA{\textit{Daimon Robotics}\\
Shenzhen, China\\
jhduan@dmrobot.com}
\and
\IEEEauthorblockN{4\textsuperscript{th} Haobo Liang}
\IEEEauthorblockA{\textit{Hong Kong Center for Construction Robotics} \\
\textit{The Hong Kong University of Science and Technology}\\
Hong Kong, China \\
hbliang@ust.hk}
\and
\IEEEauthorblockN{5\textsuperscript{th} Michael Yu Wang, \textit{IEEE Fellow}}
\IEEEauthorblockA{\textit{School of Engineering} \\
\textit{Great Bay University}\\
Dongguan, China \\ 
mywang@gbu.edu.cn}
}

\maketitle

\fancypagestyle{withfooter}{
  \renewcommand{\headrulewidth}{0pt}
  \fancyfoot[C]{\footnotesize Accepted to the Novel Approaches for Precision Agriculture and Forestry with Autonomous Robots IEEE ICRA Workshop - 2025}
}
\thispagestyle{withfooter}
\pagestyle{withfooter}

\begin{abstract}
Accurate estimation of fruit hardness is essential for automated classification and handling systems, particularly in determining fruit variety, assessing ripeness, and ensuring proper harvesting force. This study presents an innovative framework for quantitative hardness assessment utilizing vision-based tactile sensing, tailored explicitly for robotic applications in agriculture. The proposed methodology derives normal force estimation from a vision-based tactile sensor, and, based on the dynamics of this normal force, calculates the hardness. This approach offers a rapid, non-destructive evaluation through single-contact interaction. The integration of this framework into robotic systems enhances real-time adaptability of grasping forces, thereby reducing the likelihood of fruit damage. Moreover, the general applicability of this approach, through a universal criterion based on average normal force dynamics, ensures its effectiveness across a wide variety of fruit types and sizes. Extensive experimental validation conducted across different fruit types and ripeness-tracking studies demonstrates the efficacy and robustness of the framework, marking a significant advancement in the domain of automated fruit handling.

\end{abstract}

\section{Introduction}

The accurate estimation of fruit hardness is a critical factor in determining types, ripeness, and maintaining quality control during robotic picking operations. Traditional methods for assessing fruit hardness, such as mechanical probing or compression tests, often require multiple contacts or destructive testing, which are not feasible for automated systems due to their time-consuming nature and potential to damage the fruit.

Recent advancements in tactile sensing technology have significantly expanded the potential for non-destructive and efficient hardness estimation. Several studies have employed force sensors to predict object hardness. For instance, Amin et al. \cite{amin_embedded_2023} developed an embedded real-time hardness classification system for robotic grippers, while Liu et al. \cite{liu_tactclnet_2025} proposed TactCLNet, a Tactile Continual Learning Network based on generative replay for object hardness recognition. Vision-based tactile sensors, in particular, present a promising alternative by capturing detailed surface deformation information upon contact, providing real-time feedback on contact deformation. This approach has been explored by various researchers, such as Li et al. \cite{li_assessing_2024}, who assessed fruit hardness using electric gripper actuators with tactile sensors, and Yuan et al. \cite{yuan_shape-independent_2017}, who proposed a shape-independent hardness estimation method using deep learning and a GelSight tactile sensor. Nam et al. \cite{nam_softness_2024} also contributed to this field by predicting softness with a soft biomimetic optical tactile sensor. Most of these studies employ deep neural networks to process tactile images and train models for hardness prediction.

In contrast to these approaches, our proposed method utilizes three-dimensional force decomposition to analyze the average normal force dynamics during sensor-object interactions. This technique offers a robust and efficient means of estimating hardness, distinguishing it from existing methods that rely heavily on deep network architectures for tactile image processing.

The proposed framework presents several key contributions to the field of automated fruit hardness estimation. First, it enhances operational efficiency by enabling rapid hardness assessment through a single-contact interaction during the grasping process, thereby eliminating the need for additional measurement time and facilitating seamless integration into robotic systems. Second, it ensures safety and precision by utilizing a soft sensor design that provides non-destructive contact, allowing for accurate adaptation of grasping forces to minimize fruit damage. Third, the framework introduces the universal criterion based on high-resolution vision-based tactile sensors, specifically average normal force distribution dynamics, which enables generalization across diverse sizes/shapes/materials of fruits without needing variety-specific calibration. These contributions collectively advance the capabilities of robotic systems in agricultural applications, promoting more efficient and gentle handling of a wide variety of fruits.

\section{Methodology}
\subsection{Core Framework}

The principal concept underlying this research involves the utilization of a vision-based tactile sensor, which is affixed to the gripper. The sensor we adopted is the DM-Tac tactile sensor from Daimon Robotics, which can capture high-resolution tactile signals at a frequency of 120Hz and spatial resolution of 320x240. During the grasping process, the sensor can produce tactile measurements that can be decomposed into components representing normal and shear force distributions. Subsequently, we employ the dynamics of the normal force distribution to characterize the hardness of the object being grasped.

\subsubsection{Force Decomposition}

\begin{figure}[htbp]
    \centering
    \begin{subfigure}[b]{\linewidth}
        \centering
        \includegraphics[width=\linewidth]{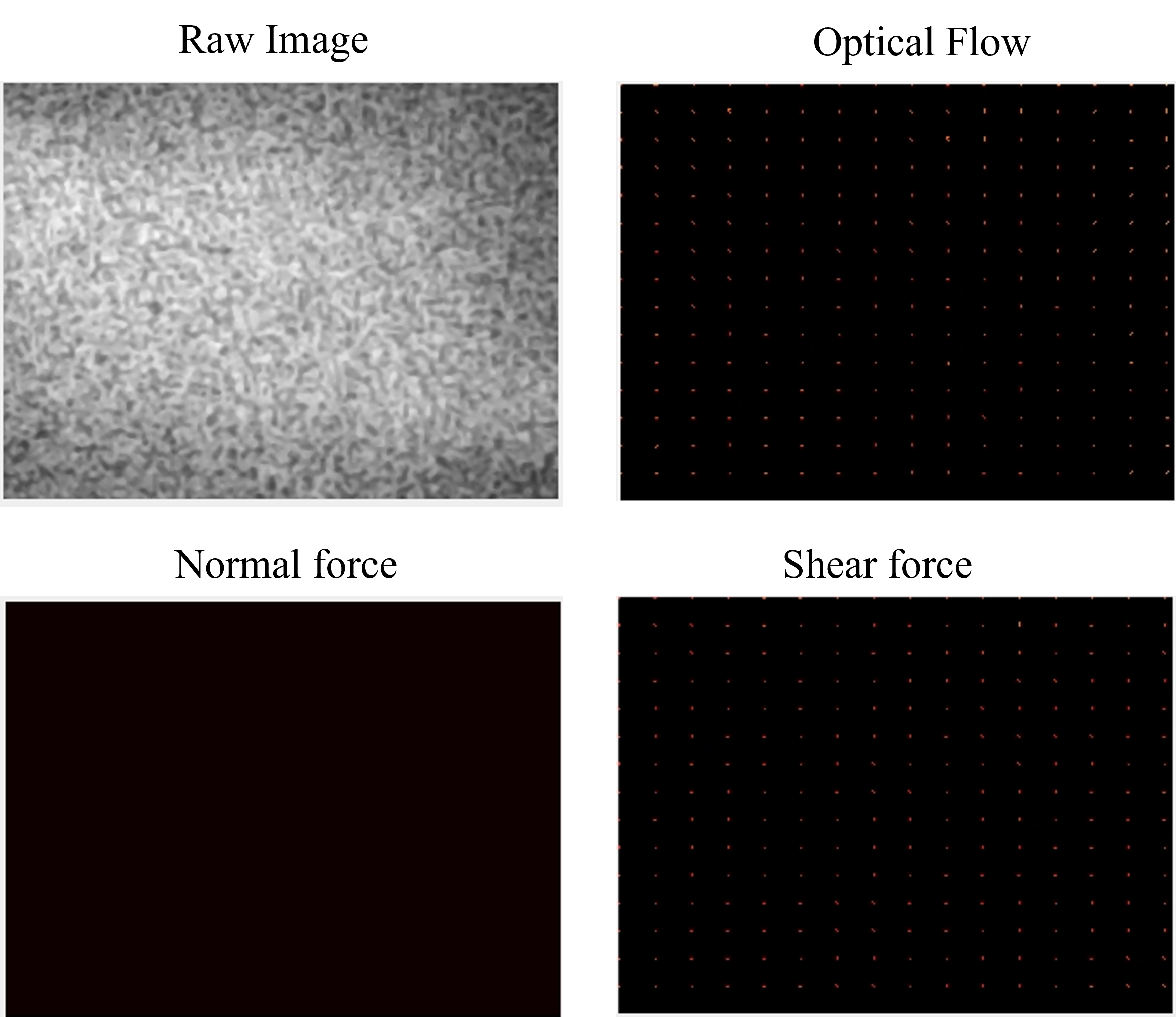}
        \caption{Original figure}
        \label{fig:original}
    \end{subfigure}
    
    \vspace{0.5cm} 
     
    \begin{subfigure}[b]{\linewidth}
        \centering
        \includegraphics[width=\linewidth]{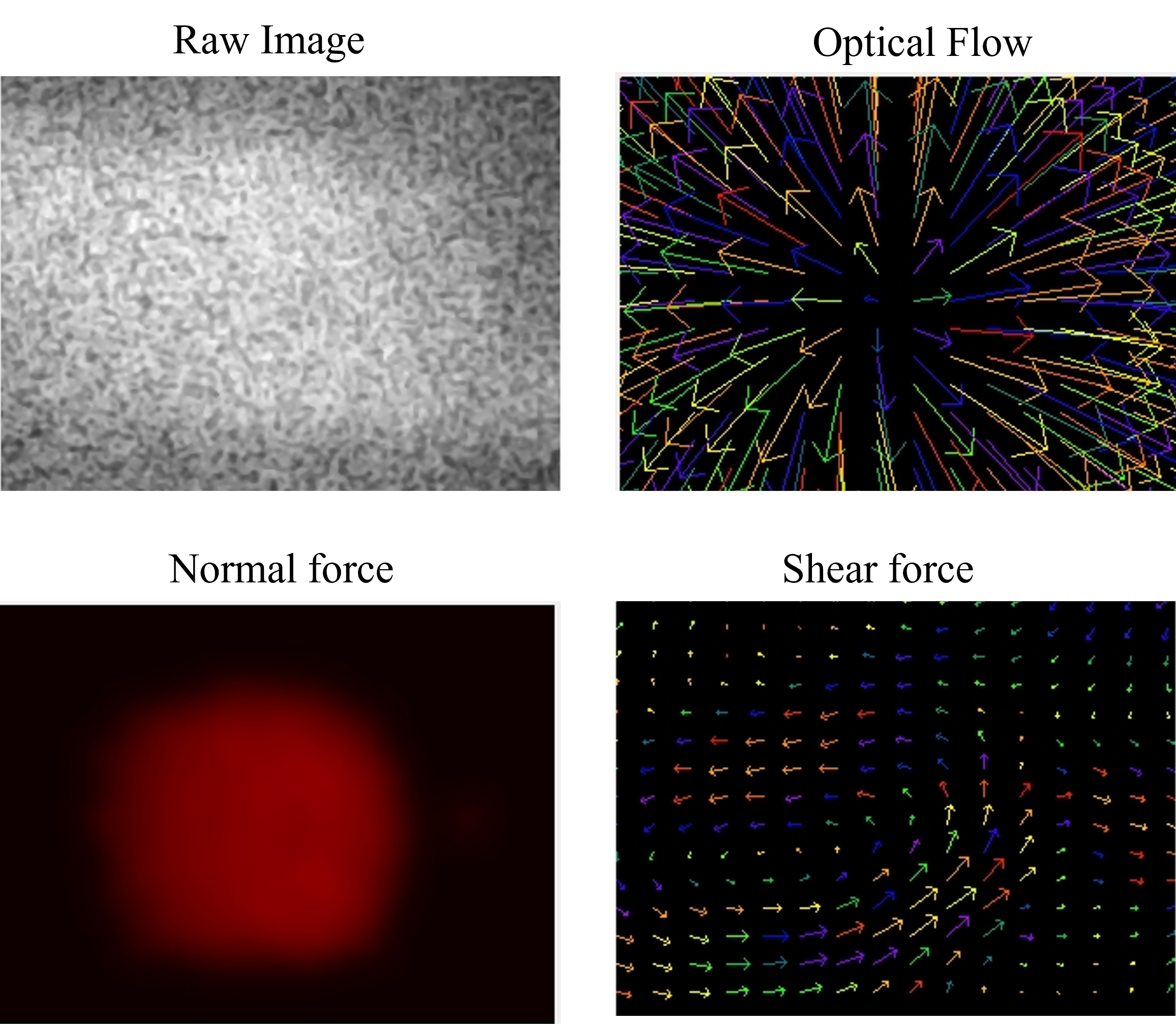}
        \caption{Deformed figure}
        \label{fig:deformed}
    \end{subfigure}
 
    \caption{Comparison of original and deformed images.}
    \label{fig:comparison}
\end{figure}

Fig. \ref{fig:comparison} illustrates the comparison between the initial state and the deformed state of the sensor surface when subjected to pressure from a cylindrical object. By analyzing the changes observed in the initial image, we can derive an optical flow. This flow is then decomposed into normal and shear force components. The methodology for force decomposition is detailed in our previous works \cite{zhang_deltact_2022, du_3d_2022}. It should be noted that the normal force in this work has no unit and only represents a relative magnitude.

\subsubsection{Hardness Estimation via Average Normal Force Dynamics}

The characterization of the normal force field concerning material hardness presents critical challenges due to continuous variations in contact area during tactile interactions. To address this, a filter function is employed for precise spatial delineation of the contact region:  
\begin{equation}
\Omega_c = \{ (x,y) \ | \ f_{\text{filter}}(\mathbf{F}_z(x,y)) > \tau \}
\end{equation}  
where \( \tau \) denotes the threshold for valid contact detection. Within \( \Omega_c \), a mathematical norm quantifies spatial variations in the normal force field $|\mathbf{F}_z|$

Two key metrics are derived from this analysis:  
\begin{itemize}
    \item Mean value: \( {\langle |\mathbf{F}_z| }\,{\rangle} \) representing average normal force
    \item Maximum value: \( {||\mathbf{F}_z||_\infty} = {\max_{(x,y)\in\Omega_c}} |\mathbf{F}_z(x,y)| \) indicating localized stress peak
\end{itemize}

This study adopts the average normal force  \( F_z = {\langle |\mathbf{F}_z| \,{\rangle}} \) as the primary hardness metric. The dynamic relationship between fruit compliance and mechanical response is established through parametric mapping:
\begin{equation}
H = g\left( \,{dF_z}/{d\delta}\,,\, {dF_z}/{dt}\,,\,\delta_{\text{max}} \right)
\label{eq:1}
\end{equation} 
where \( H \) denotes quantified hardness, \(\delta\) represents indentation depth.

The experimental protocol maintained a controlled proportionality between indentation depth $\delta$ and temporal progression $t$, establishing an invariant kinematic constraint. This operational principle enables the formulation of the hardness quantification model as: 
\begin{equation} H = g'\left( \frac{dF_z}{dt}\right) \end{equation}

In the analysis of the constant and variable rates of \(\frac{dF_z}{dt}\), we consider the following conditions:

\begin{enumerate}
    \item \textbf{Constant Rate Condition:} When $\delta_{\mathrm{max}}$ is relatively small, the change in hardness is negligible, implying a scenario where $\frac{\mathrm{d}F_z}{\mathrm{d}t} = C$. This condition suggests that the mechanical response is largely invariant to minor variations in indentation depth. 

    \item \textbf{Variable Rate Condition:} Conversely, when $\delta_{\mathrm{max}}$ is substantial, significant changes in hardness are observed, with $\frac{\mathrm{d}F_z}{\mathrm{d}t} \neq C$. This nonlinear variation is indicative of intrinsic material properties and can be leveraged for fruit attribute analysis and classification.
\end{enumerate}

This bifurcated methodological approach effectively differentiates between quasi-static experimental conditions and dynamic operational scenarios, providing insights into the fruit hardness dynamics during the grasping.

\subsection{Two Principal Criteria}

Based on the hardness estimation equation presented in Eq.\ref{eq:1}, two measurement criteria are proposed. Table \ref{tab} details the specifics of two operational modes, which are founded on two primary criteria: grasping distance and normal force.

\begin{table*}[!ht] \centering \caption{Comparative Analysis of Modes Based on Grasping Distance and Normal Force Thresholds} \begin{tabular}{|l|l|l|} \hline \textbf{Criterion} & \textbf{Grasping Distance} & \textbf{Normal Force} \\
\hline \textbf{Procedure} & \begin{tabular}[c]{@{}l@{}} 1. Initiate contact detection using sensor thresholding;\\ actuate gripper to a predetermined displacement.\\ 2. Acquire sequential average normal force values\\ and evaluate the slope as an indicator of hardness. \end{tabular} & \begin{tabular}[c]{@{}l@{}} 1. Initiate contact detection using sensor thresholding;\\ maintain closure until reaching a specified higher contact threshold,\\ then record the grasping distance.\\ 2. Derive hardness estimation from the differential in distance. \end{tabular} \\ \hline \textbf{Use Case} & \begin{tabular}[c]{@{}l@{}} Applications with precision requirements necessitating\\ detailed ripeness assessment. \end{tabular} & \begin{tabular}[c]{@{}l@{}} Tasks prioritized by time constraints such as bulk sorting\\ or automated pick-and-place operations. \end{tabular} \\ \hline \end{tabular} \label{tab}\end{table*}

In the domain of tactile sensing for fruit hardness estimation, the request for precision and efficiency frequently results in a compromise, as achieving both concurrently poses a formidable challenge. High precision demands slower operational speeds and deeper grasp engagements for the acquisition of comprehensive data, whereas efficiency necessitates expedited operations and minimal contact deformation.

Criteria based on normal force, typically employed by humans when assessing hardness, involve using a finger to apply pressure until a threshold is reached, at which point the deformation distance is used to perceive hardness. This method generally encompasses a constant rate condition. When the maximum deformation, $\delta_{\mathrm{max}}$, is relatively small, the slope $\frac{\mathrm{d}F_z}{\mathrm{d}t} = C$ can be directly correlated with hardness, which is essential for classifying fruit types.

\begin{figure}
    \centering
    \includegraphics[width=1\linewidth]{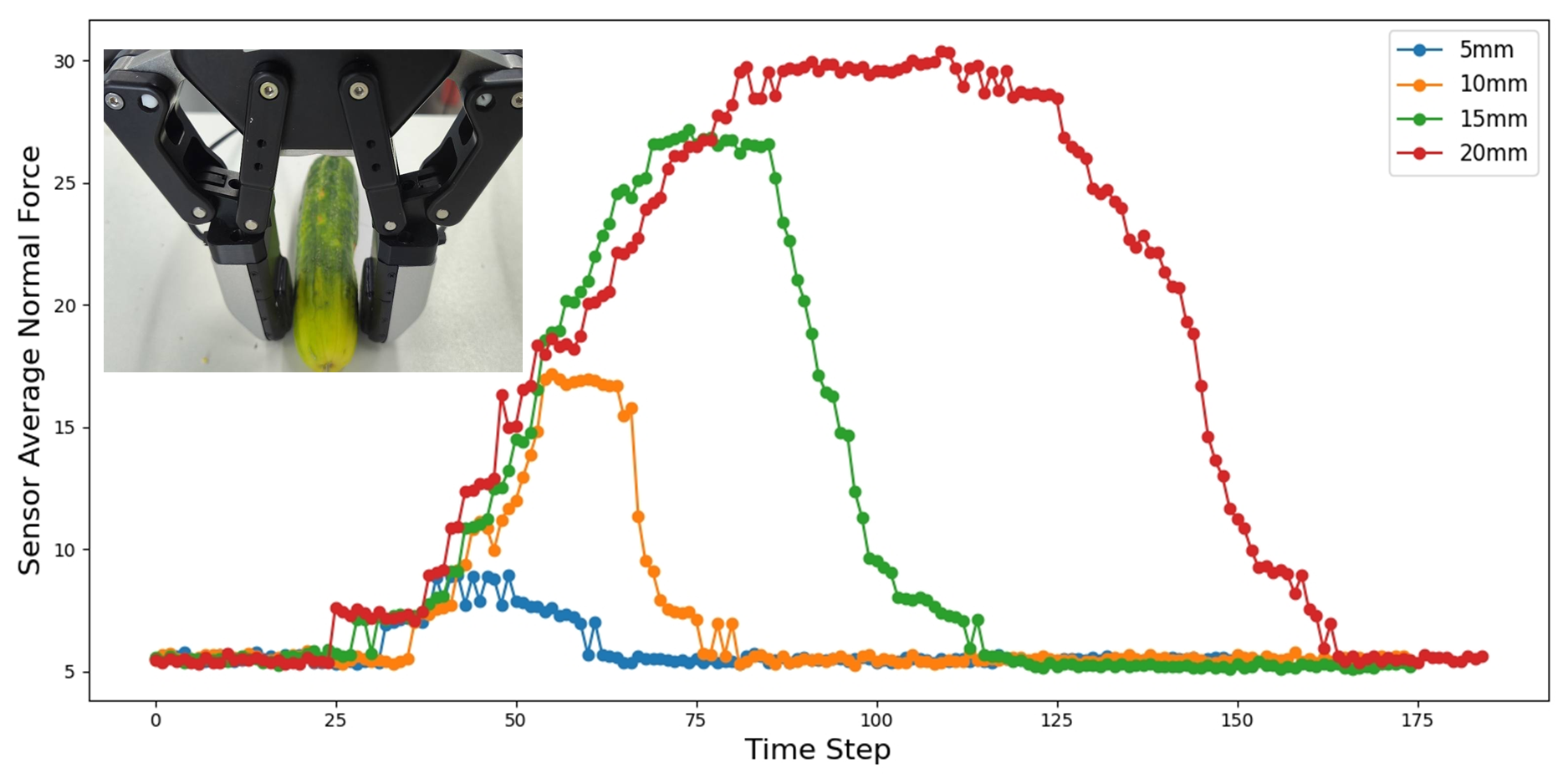}
    \caption{Different grasp distance for old cucumber (very soft).}
    \label{fig:oldcucumber}
\end{figure}

In contrast, criteria based on distance may lead to a variable rate condition, characterized by the inequality $\frac{\mathrm{d}F_z}{\mathrm{d}t} \neq C$, which suggests nonlinear variations. Given the high frame rate capability of the sensor, operating at 120 frames per second, it is adept at capturing the fluctuations in hardness throughout the grasping process. The initial slope, denoted as $C$, can be employed to quantify hardness, while the second derivative of force, $\frac{\mathrm{d}^2 F_z}{\mathrm{d}t^2}$, aids in the assessment of fruit characteristics, including ripeness. Fig. \ref{fig:oldcucumber} illustrates the variations in distance (ranging from 5 mm to 20 mm) for an old cucumber, which is characterized by a very soft texture. It is observed that the slope $C$ exhibits only slight changes. Beyond 20 mm, there is a marked escalation in force. During the experiment, the force measured at 25 mm was over three times greater than that at 20 mm, leading to a pronounced increase in the slope.

\section{Experimental Validation}
\subsection{Benchmarking Across Fruit Varieties}

\begin{figure}
    \centering
    \includegraphics[width=1\linewidth]{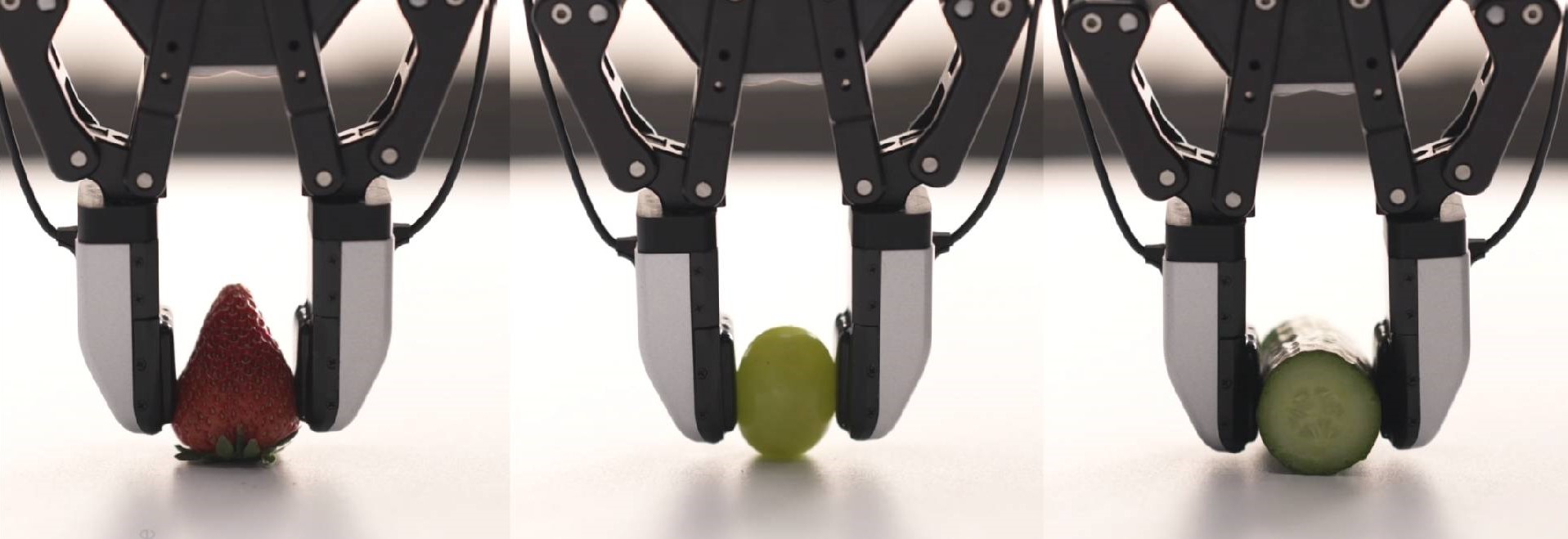}
    \caption{Three kinds of fruits: cucumbers, strawberries, and grapes. }
    \label{fig:3fruit}
\end{figure}

\begin{figure}
    \centering
    \includegraphics[width=1\linewidth]{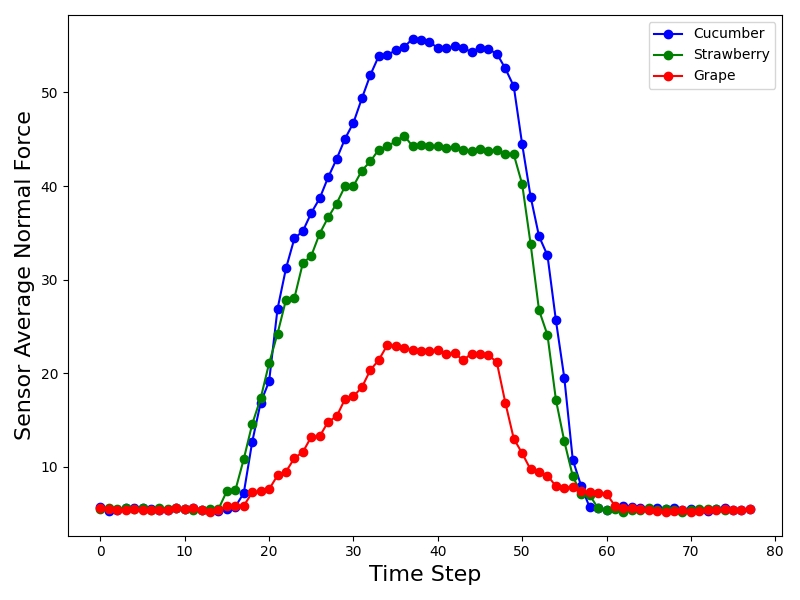}
    \caption{Continue grasping with the same grasp distance. The normal force value only represents relative magnitude.}
    \label{fig:1}
\end{figure}

\begin{figure}
    \centering
    \includegraphics[width=1\linewidth]{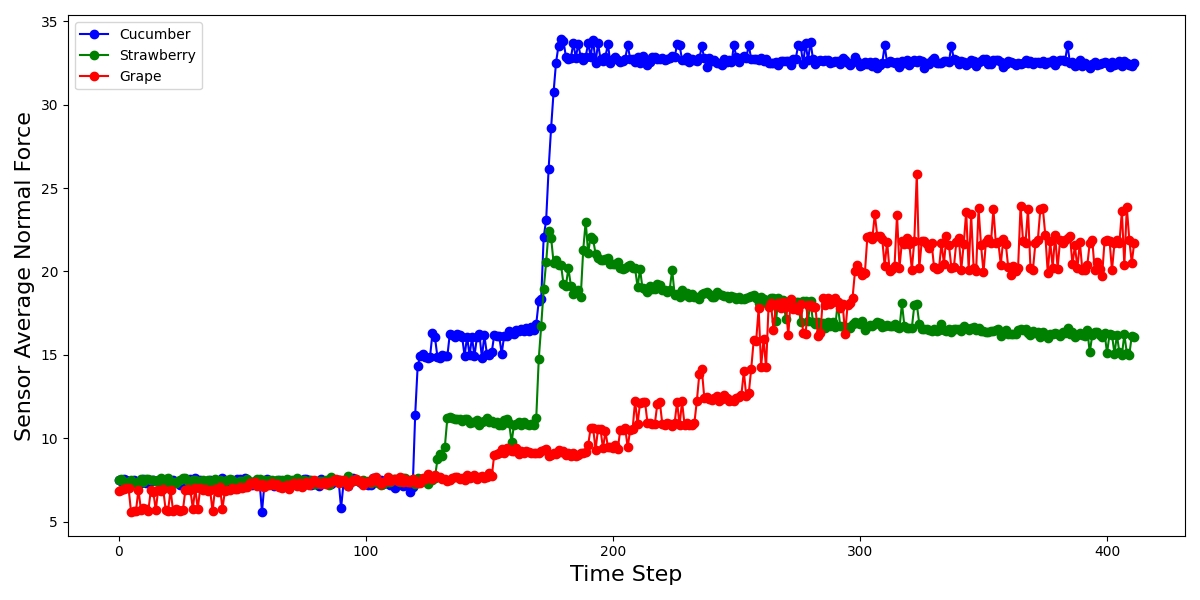}
    \caption{Step grasping based on the normal force threshold.}
    \label{fig:2}
\end{figure}

Systematic testing of various fruits along the soft-to-hard spectrum, including cucumbers, strawberries, and grapes, confirms the applicability of the established criteria. Fig. \ref{fig:3fruit} illustrates the experimental setup, while Fig. \ref{fig:1} depicts continuous grasping at a consistent grasp distance. Harder fruits exhibit a significant increase in force over time, resulting in a higher slope. In Fig. \ref{fig:2}, a step-wise grasping approach based on a normal force threshold is presented. Due to the limitations of the gripper, it is challenging to halt the grasping mechanism at an exact value; consequently, we opted for a step-wise grasping method with small increments of 1.5 mm. The force threshold for this experiment was set at 20. Notably, the hardest fruit, the cucumber, reached the threshold more rapidly and generated a greater force, while the softer fruits required more steps to achieve the same threshold. Both criteria demonstrated effectiveness in classifying the different fruits.

\subsection{Ripeness Tracking Case Study – Mango \& Kiwi fruit}

\begin{figure}[ht] \centering \includegraphics[width=1\linewidth]{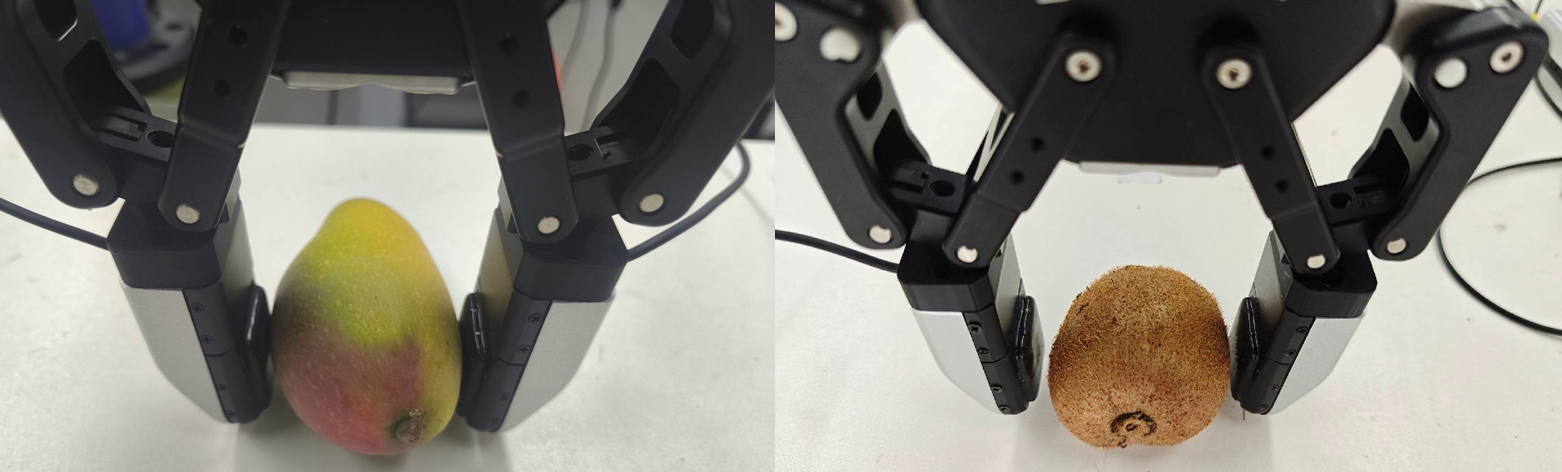} \caption{Hardness measurement using sensor-integrated gripper for non-destructive fruit ripeness evaluation}
\label{fig:ggg} \end{figure}

\begin{figure}[ht]
    \centering
    \begin{subfigure}[b]{0.9\linewidth}
        \centering
        \includegraphics[width=\linewidth]{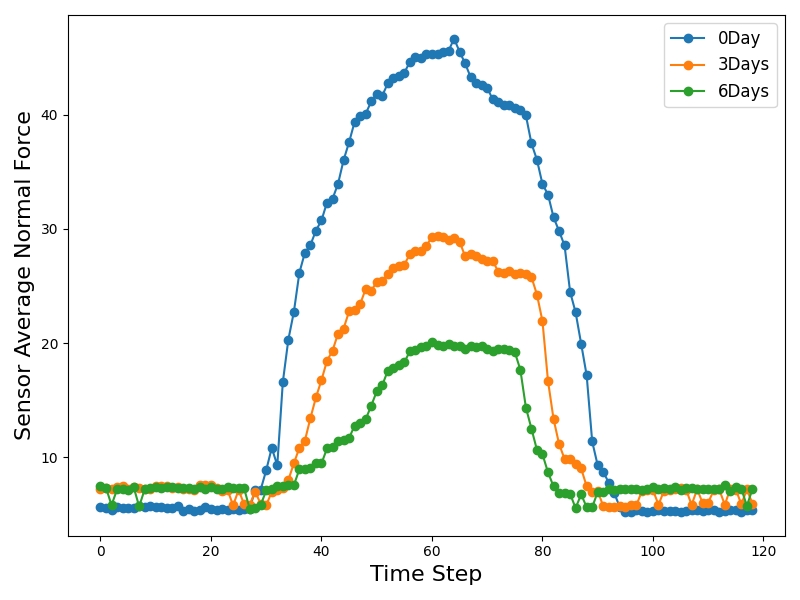}
        \caption{Mango}
        \label{fig:mango}
    \end{subfigure}
    
    \vspace{0.5cm} 
    
    \begin{subfigure}[b]{0.9\linewidth}
        \centering
        \includegraphics[width=\linewidth]{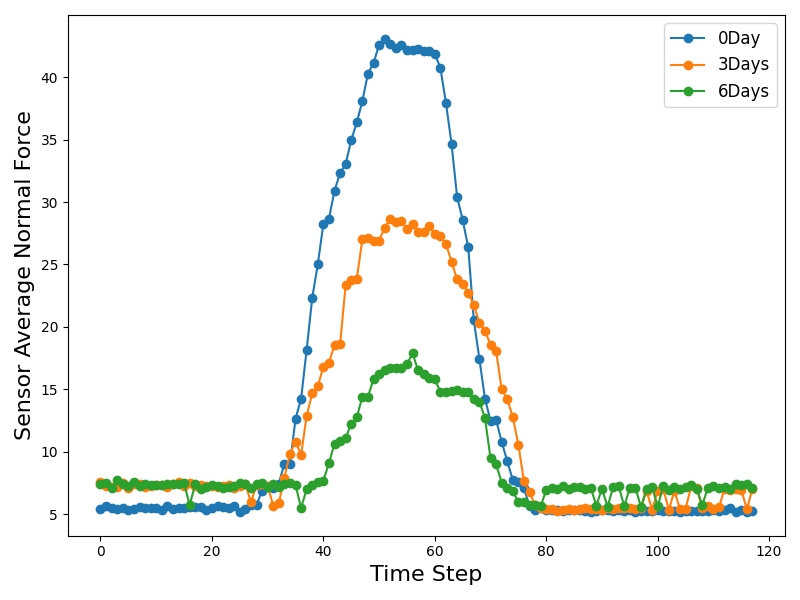}
        \caption{Kiwi}
        \label{fig:kiwi}
    \end{subfigure}
    
    \caption{Biomechanical ripening trajectories of mango (a) and kiwi (b) specimens}
    \label{fig:combined}
\end{figure}

Longitudinal monitoring conducted across multiple diurnal cycles revealed the methodology's capacity to detect subtle textural modifications associated with post-harvest physiological changes. Fig. \ref{fig:ggg} illustrates the experimental configuration employing a sensor-integrated gripper apparatus for quantitative fruit firmness assessment. Subsequent analyses presented in Fig. \ref{fig:combined} mango(a) and kiwi fruit (b) demonstrate similar ripening progression patterns.

The datasets exhibit analogous temporal trends characterized by distinct biomechanical phase transitions. Initial measurements (0 Day) displayed rapid normal force escalation under equivalent compression distances, achieving maximal force thresholds within equal closure distance. Progressive temporal evolution manifested through decreasing first-derivative values $\frac{\mathrm{d}F_z}{\mathrm{d}t} = C$, accompanied by significant reductions in peak normal force magnitudes. These parametric variations correlate strongly with pericarp structural degradation phenomena, indicative of progressive tissue softening concomitant with advancing maturation stages.

\subsection{Grasp Optimization Trials}

\begin{figure}
    \centering
    \includegraphics[width=1\linewidth]{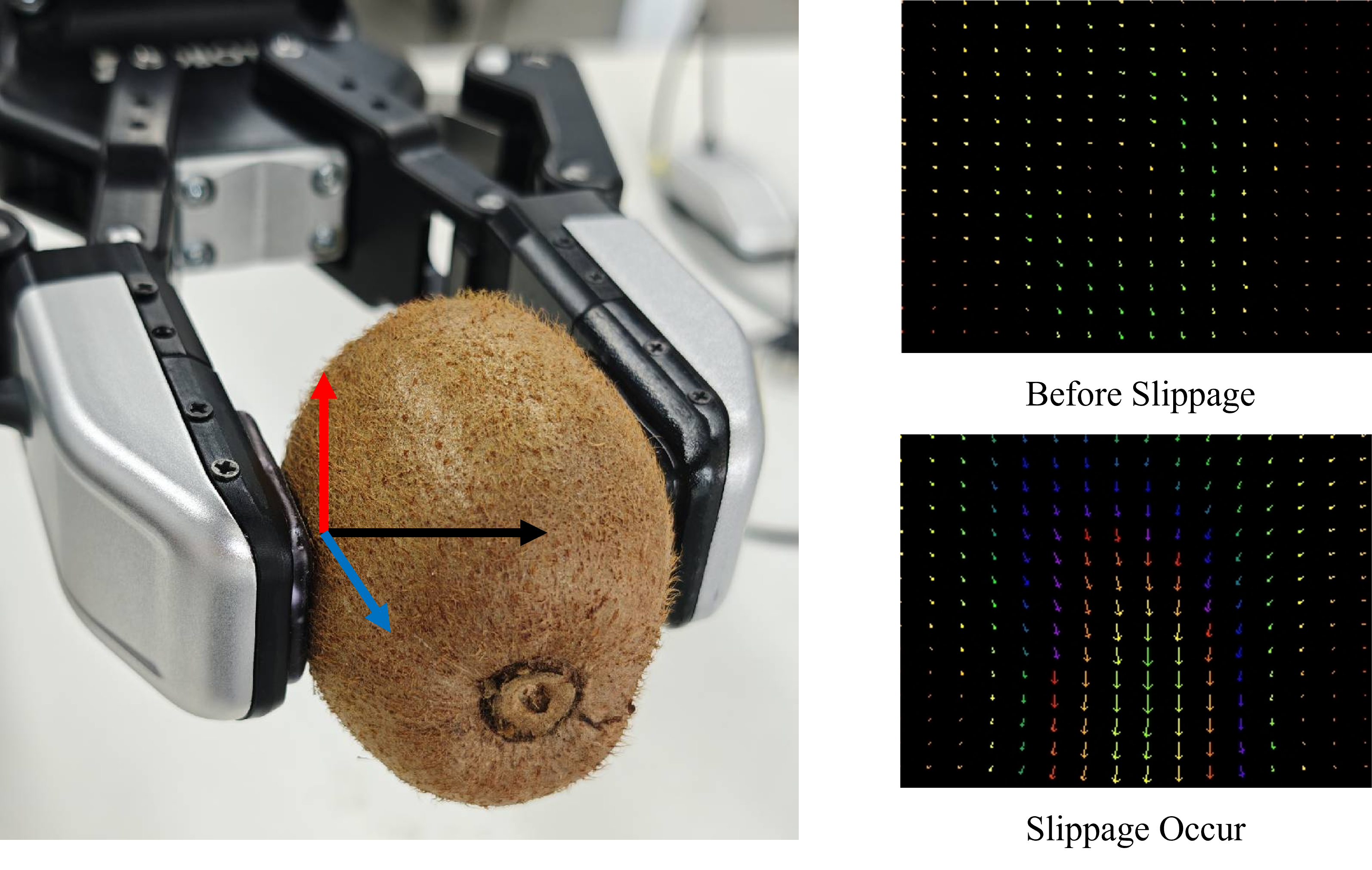}
    \caption{Accurate grasp visualization. The shear force changed upon the grasped object's status.}
    \label{fig:Accurategrasp}
\end{figure}

\begin{figure}
    \centering
    \includegraphics[width=1\linewidth]{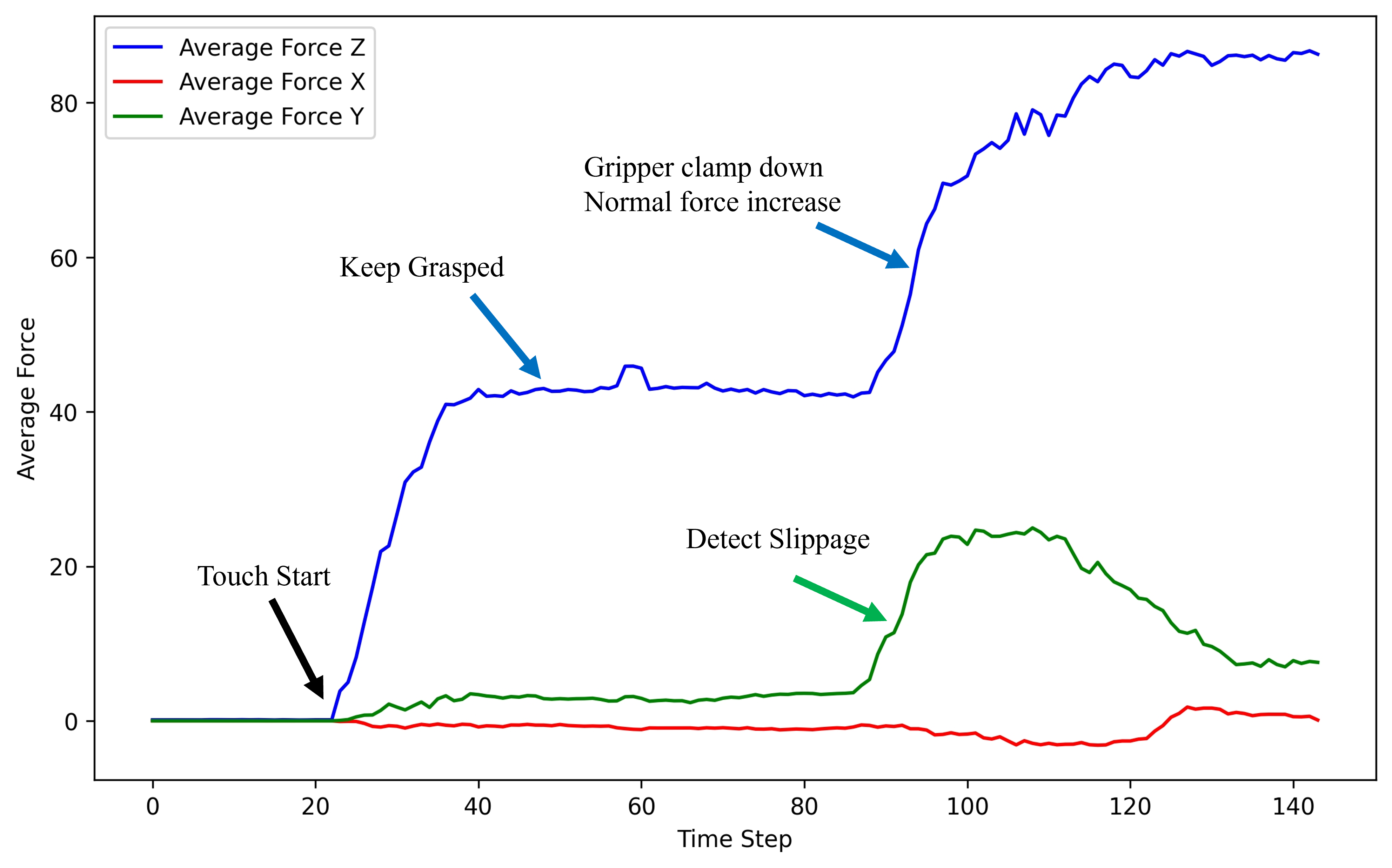}
    \caption{Slippage detection. The sensor can measure three-dimensional forces, enabling it to detect slippage. Upon detection of slippage, the gripper adapts by adjusting its closure.}
    \label{fig:3f}
\end{figure}

The vision-based tactile sensor functions at an elevated frame rate of 120 frames per second. However, the existing grippers are unable to attain such a high frequency of control. In contrast to conventional sensors, the integration of the gripper with the vision-based tactile sensor facilitates enhanced precision in grasping capabilities. For delicate fruits that are prone to damage during handling, the sensor provides real-time feedback to the gripper, allowing for accurate determination of the force required to halt the grasping process, as shown in Fig. \ref {fig:Accurategrasp}. Additionally, the sensor is capable of measuring three-dimensional forces, which enables it to detect slippage. In Fig.\ref{fig:3f}, upon detection of slippage, the gripper adapts by adjusting its closure.

\section{Conclusion}

This study introduces a comprehensive framework for assessing fruit hardness using tactile perception, offering notable enhancements in efficiency, safety, and adaptability for robotic applications. By leveraging the unique capabilities of vision-based tactile sensors, the proposed approach facilitates rapid and non-destructive evaluation of fruit hardness, making it highly suitable for integration into automated agricultural systems. Future research will concentrate on the utilization of sensor data, as this study exclusively employs normal force data. By incorporating 3D force measurements, it is possible to enhance the estimation of fruit status. Additionally, efforts will be directed toward extending the framework's applicability to a wider array of tasks in agricultural robotics.

\end{document}